\pdfoutput=1

\documentclass[11pt]{article}

\usepackage[final]{acl}

\usepackage{times}
\usepackage{latexsym}
\usepackage{makecell}
\usepackage{enumitem}
\usepackage[T1]{fontenc}

\usepackage[utf8]{inputenc}

\usepackage{microtype}

\usepackage{inconsolata}

\usepackage{graphicx}

\usepackage{hyperref}
\usepackage{url}
\usepackage{booktabs}       
\usepackage{amsfonts}       
\usepackage{nicefrac}       
\usepackage{multirow}
\usepackage{amsmath}
\usepackage{colortbl}
\usepackage{algorithm}
\usepackage{algpseudocode}
\usepackage{arydshln}
\definecolor{LightCyan}{rgb}{0.93,0.95,1}
\newcolumntype{b}{>{\columncolor{LightCyan}}c}
\usepackage{minitoc} 

\title{Agentic Reasoning: A Streamlined Framework for Enhancing LLM Reasoning with Agentic Tools}

\author{
 \textbf{Junde Wu\textsuperscript{1,2}},
 \textbf{Jiayuan Zhu\textsuperscript{1}},
 \textbf{Yuyuan Liu\textsuperscript{1}},
 \textbf{Min Xu\textsuperscript{3,4}},
 \textbf{Yueming Jin\textsuperscript{2}\thanks{Corresponding author}}
\\
 \textsuperscript{1}University of Oxford,
 \textsuperscript{2}National University of Singapore, \\
 \textsuperscript{3}Carnegie Mellon University,
 \textsuperscript{4}MBZUAI
\\
}

\begin{document}
\maketitle
\begin{abstract}
We introduce Agentic Reasoning, a framework that enhances large language model (LLM) reasoning by integrating external tool-using agents. Agentic Reasoning dynamically leverages web search, code execution, and structured memory to address complex problems requiring deep research. A key innovation in our framework is the Mind-Map agent, which constructs a structured knowledge graph to store reasoning context and track logical relationships, ensuring coherence in long reasoning chains with extensive tool usage. Additionally, we conduct a comprehensive exploration of the Web-Search agent, leading to a highly effective search mechanism that surpasses all prior approaches. When deployed on DeepSeek-R1, our method achieves a new state-of-the-art (SOTA) among public models and delivers performance comparable to OpenAI Deep Research, the leading proprietary model in this domain. Extensive ablation studies validate the optimal selection of agentic tools and confirm the effectiveness of our Mind-Map and Web-Search agents in enhancing LLM reasoning. The code is at: \url{https://github.com/theworldofagents/Agentic-Reasoning}.
\end{abstract}

\addtocontents{toc}{\protect\setcounter{tocdepth}{-10}} 

\begin{table}[h]
    \centering
    \resizebox{0.95\hsize}{!}{
    \begin{tabular}{l c}
        \toprule
        \textbf{Model} & \textbf{Accuracy (\%)} \\
        \midrule
        GPT-4o$^\dagger$ & 3.3 \\
        Grok-2$^\dagger$ & 3.8 \\
        Claude 3.5 Sonnet$^\dagger$ & 4.3 \\
        Gemini Thinking$^\dagger$ & 6.2 \\
        OpenAI o1$^\dagger$ & 9.1 \\
        \textbf{DeepSeek-R1} & \textbf{9.4} \\
        OpenAI o3-mini (medium)$^\dagger$ & 10.5 \\
        OpenAI o3-mini (high)$^\dagger$ & 13.0 \\
        \textbf{Agentic Reasoning w/ R1} & \textbf{23.8} (\textcolor{red}{+14.4}) \\
        \textit{Perplexity deep research} $^\dagger$ & \textit{21.1} \\
        \textit{OpenAI deep research} $^\dagger$ & \textit{26.6} \\
        \bottomrule
        $^\dagger$ denotes proprietary models.
    \end{tabular}}
    \caption{On Humanity's Last Exam, we achieved a remarkable 23.8\% with DeepSeek-R1, marking a 14.4\% improvement over the base model. This narrows the gap to the proprietary OpenAI Deep Research to just 2.8\%, which depends on a stronger internal reasoning model. }
    \label{tab:model_accuracy}
\end{table}

\section{Introduction}
Recently, large reasoning models, such as OpenAI’s o1 \cite{jaech2024openai}, Qwen-QwQ \cite{teamqwq}, and DeepSeek-R1 \cite{team2024deepseek}, have demonstrated impressive stepwise reasoning capabilities over long sequences through large-scale reinforcement learning. These advancements provide promising solutions to complex reasoning tasks \cite{wei2022chain, lewkowycz2022solving, noauthor_learning_nodate} and have inspired foundational efforts to replicate o1-like reasoning patterns across a broader range of models \cite{qin2024o1, huang2024o1, zhang2024llama}. It is recently revealed by DeepSeek-R1 that applying rule-based outcome rewards during training, such as evaluating whether a piece of code executes successfully, could yield remarkable reasoning capabilities equaling o1-level math and coding performance.

Although current reasoning methods excel in structured domains like math and code—where outcomes are easily verifiable—applying these techniques to less structured or knowledge-intensive tasks remains a significant challenge. As mentioned in DeepSeek-R1 \cite{team2024deepseek}, not all problems benefit from formal reasoning approaches. Many fields, such as social sciences, ethics, or experiential disciplines, rely on abstract concepts, conventional wisdom, factual verification, understanding complex logical relationships, or moral reasoning. When models attempt to impose math- or coding-style reasoning onto such areas, they often produce flawed or overly rigid results. Developing approaches that account for these unique requirements is essential for advancing the applicability of reasoning model beyond their current domains.

Deep, thoughtful answers to open-ended questions often require extensive research, repeated verification, information retrieval, computational analysis, and the organization of complex logical relationships—steps fundamental to human reasoning. In this process, humans rely heavily on external tools, such as internet searches for gathering information, computational tools for quantitative analysis, or whiteboards and mind maps for organizing thoughts. This raises an intriguing question: can reasoning LLMs similarly leverage external tools to enhance their reasoning and tackle intensive knowledge work across diverse domains?

Previous efforts have attempted to integrate search or retrieval-augmented generation (RAG) into the reasoning process \cite{shao2024assisting, khaliq2024ragar, islam2024open, li2025search}, with notable examples including Gemini's and OpenAI's Deep Research. However, these models are proprietary, and their exact methodologies remain undisclosed. In contrast, open-source models primarily focus on retrieval and web-search integration during reasoning but still exhibit a notable performance gap compared to their closed-source counterparts.

We introduce Agentic Reasoning, a framework that enhances reasoning by integrating external LLM-based agents as tools. This approach allows LLMs to delegate specific tasks to auxiliary agents during the reasoning process, thereby improving their overall problem-solving capabilities. Through extensive experimentation with integrating various agents into the reasoning process, we identified three essential agents that prove highly effective for general reasoning across diverse problems. The Web-Search agent, which retrieves relevant information from the internet to supplement the model's knowledge. The Code agent, capable of performing computational analyses and coding tasks to support quantitative reasoning. Finally, the memory agent, which we call Mind-Map, constructs knowledge graphs based on the reasoning context, enabling the organization of complex logical relationships in a manner similar to human mind mapping. Together, these agents enhance the model's ability to tackle complex problems and do deep research with greater efficiency and precision.

We evaluated our model on general knowledge-intensive benchmarks requiring complex reasoning capabilities, categorized into two key areas: (1) solving expert-level questions and (2) conducting deep research on real-world expert-level tasks. For expert-level questions, we evaluate the model on Humanity’s Last Exam \cite{phan2025humanitysexam}, a recently released benchmark assessing AI performance across a broad range of subjects. As shown in Table \ref{tab:model_accuracy}, we achieves a new high of 23.8\% accuracy, marking a 14.4\% improvement over the raw model and narrowing the open-source vs. closed-source gap to just 2.8\% compared to the proprietary OpenAI Deep Research, which benefits from a stronger internal reasoning model. For real-world expert-level tasks, Agentic Reasoning was assessed by domain experts, who found that it effectively automated complex manual investigation. This underscores its potential to streamline labor-intensive processes and boost productivity in knowledge-intensive domains.

In brief, our contribution can be concluded as:
\begin{itemize}
\item  We propose Agentic Reasoning, a streamlined framework that enhances reasoning by integrating external LLM-based agentic tools. We experimentally identify web-search, coding, and Mind-Map agents as three universally effective tools.
\item We explore the design of the Web-Search agent and identify a strategy that outperforms previous search or RAG approaches.
\item We develop a knowledge-graph-based Mind-Map to assist reasoning, improving the model’s ability to handle complex logic and maintain coherence in long reasoning chains.
\item We evaluate our approach on expert-level problem-solving and deep research tasks, achieving new SOTA results across several benchmarks and surpassing prior methods in human evaluations.
\end{itemize}

\begin{figure*}[t]
    \begin{center}
    \includegraphics[width=0.95\textwidth]{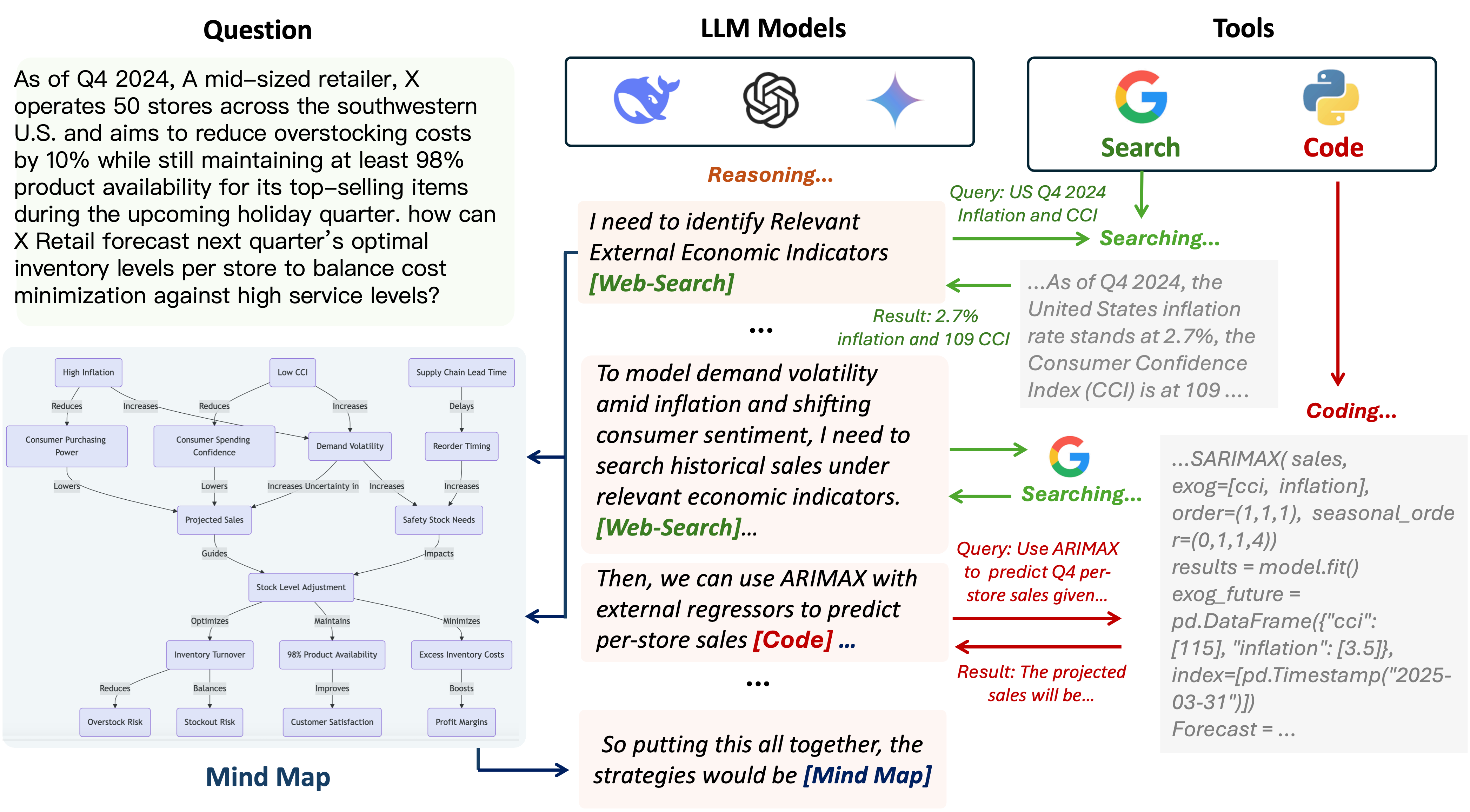}
    \end{center}
    \caption{The overall workflow of Agentic Reasoning. Given a question, the reasoning LLM can invoke the Web-Search agent to retrieve external information, the Coding agent to perform quantitative computations, and the Mind-Map agent to structurally memorize the reasoning context, to provide a comprehensive solution.}\label{fig1}
\end{figure*}

\section{Method}

\subsection{Agentic Reasoning Pipeline}
Our core idea is to enhance LLM reasoning by integrating external LLM-based agents into the process. During reasoning, the model can call these agents as tools to assist in problem-solving while maintaining a structured memory to store its reasoning context. In the overall process, we deploy a Web-Search agent and a Code agent as problem-solving tools, along with a knowledge-graph agent, called Mind-Map, to serve as memory.

Specifically, the reasoning LLM can dynamically determine when to call external agentic tools during its reasoning process. As shown in Figure \ref{fig1}, when needed, it embeds specialized tokens into its reasoning sequence, categorizing them as web-search tokens, coding tokens, or Mind-Map calling tokens. Alongside these tokens, the reasoning LLM generates a query as a message to the external agents.

Upon detecting such a token, the reasoning process temporarily halts to extract the query and its reasoning context. These queries are then dispatched to the corresponding external agents. The agents would consider both the received query and the reasoning context to ensure the most pertinent results are returned to the main reasoning chain. These results are then reintegrated into the reasoning chain, allowing the model to continue its inference with an updated knowledge. This iterative retrieval-and-reasoning cycle continues as needed, enabling the model to dynamically refine its reasoning until it reaches a fully reasoned final answer.

\subsection{Mind-Map Agent}
We construct a Mind-Map to store and structure the real-time reasoning context of the reasoning model. This Mind-Map is built by transforming raw reasoning chains into a structured knowledge graph. Specifically, we use a graph-construction LLM to extract entities from the reasoning chain and identify semantic relationships between related entities, following a process similar to that used in GraphRAG \cite{edge2024local}.

The Mind-Map serves two primary functions. First, it clusters reasoning context into distinct groups and summarizes each of them. This is achieved by applying community clustering \cite{traag2019louvain} on the knowledge graph and using an LLM to generate concise summaries for each group. Second, the knowledge graph can be queried with specific questions, such as “Who was Jason's maternal great-grandfather?” Using standard RAG on the knowledge graph \cite{edge2024local}, we retrieve and return the relevant information to response the query.

These functions integrate the Mind-Map into two key aspects of the Agentic Reasoning process. First, it provides reasoning context to external tools, enabling them to generate more context-aware responses. The context is generated by synthesizing the summaries of each clustered group, performed by an LLM. Additionally, when the reasoning model encounters uncertainty or loses track in an extended reasoning process, it can query the Mind-Map as an external memory to retrieve relevant information and continue reasoning seamlessly. This ensures the model maintains a long reasoning chain across multiple breakdown tasks and tool calls without missing critical information.

\subsection{Web-Search agent}
A search agent is invoked to retrieve the most relevant documents from the web. It consists of four key components: query breakdown, a search service, a re-ranking service, and RAG.

When the reasoning model generates a web-search query, it is sent to the Web-Search agent, which first reorganizes it into one or more search-optimized queries suitable for search engines like Google or Bing. The process involves sending the LLM the original query along with the reasoning context retrieved from the Mind-Map, prompting it to generate suitable refined search queries. For example, given the original query "Search the external economic indicators" and the context "We are looking for the optimal investing strategy for a retailer in the U.S. in Q4 2024", the Web-Search agent would break it down into more specific queries such as "U.S. Q4 2024 inflation rate" and "U.S. Q4 2024 CCI". These queries are then sequentially sent to the search engine, which returns related web pages.

After we retrieved the web pages for each refined query, we apply a re-ranking model to rank web pages based on their alignment with the original query and context. The average relevance score of the top 10 pages is computed, and if it falls below a predefined threshold, the Web-Search agent will iterate back to the last step and further refine the search query.

Once reranking is complete, web pages with relevance scores above the threshold are stored, and RAG is applied on them to extract meaningful insights. Each refined query undergoes RAG to generate a natural language response. Finally, an LLM synthesizes these responses into a cohesive final snippet, based on both the original query and reasoning context. This processed snippet is then integrated into the main reasoning process, ensuring that external insights enhance logical flow without causing disruption.
\subsection{Coding Agent}
Instead of prompting the reasoning model to generate code directly, we find it more efficient to delegate coding tasks to a specialized coding LLM. The reasoning model sends the relevant context and query message to the coding LLM, which then writes the required code, executes it via a compiler, and returns the results. This approach ensures that the reasoning model remains focused on its core reasoning process without being disrupted by coding tasks, allowing for longer and more coherent reasoning chains. Specifically, we format the coding request as follows:
"Write code to perform <code message from reasoning model> given the context <reasoning context from Mind-Map> to answer the query <user query>." The coding LLM is instructed to always return its output in natural language, ensuring seamless integration with the reasoning model.

\begin{figure}[h]
    \begin{center}
    \vspace{-20pt}
    \includegraphics[width=0.4\textwidth]{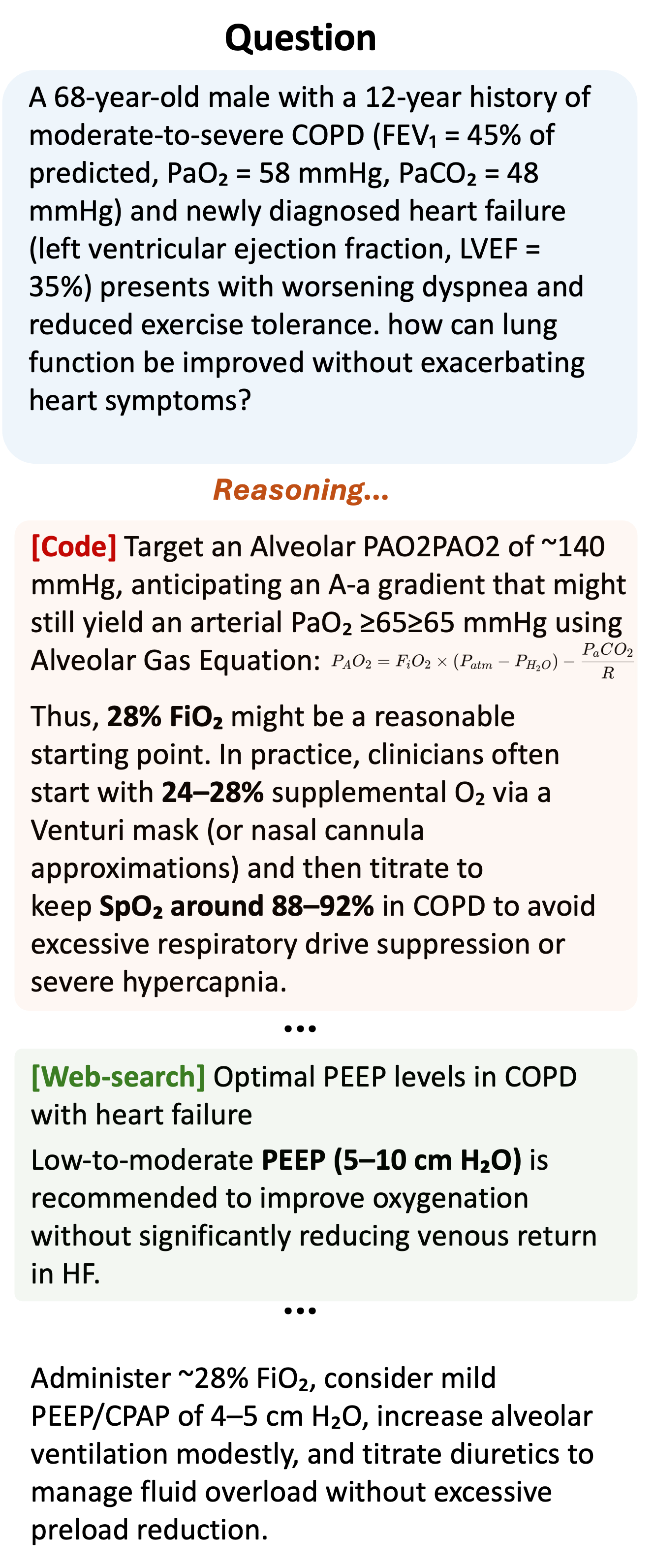}
    \end{center}
    \vspace{-10pt}
    \caption{Case study on a complex medical decision-making problem.}\label{medex}
\end{figure}

\section{Experiments}
\begin{table}[ht]
\centering
\caption{Performance comparison on GPQA dataset across Physics, Chemistry, and Biology.}
\resizebox{0.95\hsize}{!}{
\begin{tabular}{lcccc}
\hline
\textbf{Method} & \textbf{Phy.} & \textbf{Chem.} & \textbf{Bio.} & \textbf{All}  \\
\hline
\multicolumn{4}{l}{\textit{Direct Reasoning}} \\
QwQ-32B & 75.6 & 39.8 & 68.4 &58.1 \\
Llama3.3-70B & 54.7 & 31.2 & 52.6 &43.4 \\
DeepSeek-R1 & 86.8 & 56.1 & 63.8 & 71.5  \\
GPT-4o$^\dagger$ & 59.5 & 40.2 & 61.6 &50.0  \\
o1$^\dagger$ & 92.8 & 64.7 & 69.2 &78.0  \\
o3-mini-low$^\dagger$ & - & - & - &70.6  \\
o3-mini-mid$^\dagger$ & - & - & - &76.8   \\
o3-mini-high$^\dagger$ & - & - & - &79.7  \\
\hline
\multicolumn{4}{l}{\textit{Retrieve/Search in Reasoning}} \\
RAgent w/QwQ-32B & 76.7 & 46.2 & 68.4 & 61.6  \\
RAgent w/DeepSeek-R1 & 87.7 & 58.2 & 65.7 & 72.9 \\
SearchO1 w/QwQ-32B & 77.9 & 47.3 & 78.9 & 63.6 \\
SearchO1 w/DeepSeek-R1 & 90.2 & 61.3 & 71.4  & 74.6 \\
\hline
\multicolumn{4}{l}{\textit{Agentic Reasoning}} \\
Ours w/QwQ-32B & \textbf{88.1} & \textbf{58.3} & \textbf{79.6} & \textbf{69.7} \\
Ours w/DeepSeekR1 & \textbf{94.5} & \textbf{73.7} & \textbf{80.5} & \textbf{81.2} \\ 
\hline
\end{tabular}}\label{gpqa1}
\end{table}

\subsection{Implementation Details}
In our experiments, we use DeepSeek-R1 as the primary reasoning models by default. For the Web-Search agent, query breakdown and RAG are handled by DeepSeek-V3 \cite{liu2024deepseek}. We use a maximum of 32,768 tokens, temperature of 0.7, top\_p of 0.8, top\_k of 20, and a repetition penalty of 1.05 across all models for generation. We use Bing as the search engine, retrieving the top 20 most relevant pages. The re-ranking model is Cohere Rerank 3.5, with a top-10 average relevance score threshold of 0.7 to determine if iterative query refinement is needed, allowing a maximum of three iterations. Additionally, web pages with a relevance score above 0.7 are selected for RAG processing. For the Mind-Map agent, both knowledge graph construction and Graph-RAG retrieval are also performed using DeepSeek-V3. For the coding agent, we use claude-3.5-sonnet to generate code and Python 3.11 for execution. We report pass\@1 results by default. 

\subsection{Solving Expert-level Problems}
Agentic Reasoning model is able to call external tools in its reasoning to solve expert-level problems, except Humanity’s Last Exam we previously mentioned, we further evaluate it on two datasets: GPQA dataset \cite{rein2023gpqa}, a PhD-level multiple-choice science QA benchmark, and GAIA \cite{mialon2023gaia}, a benchmark for AI agents that requires a set of abilities such as reasoning, web browsing, and tool-use proficiency. 

As shown in Table \ref{gpqa1}, applying Agentic Reasoning to a strong reasoning model like DeepSeek-R1 achieves a new SOTA, surpassing even the best proprietary model, o3-mini-high. Compared to the base model DeepSeek-R1, our method boosts overall performance by nearly 10\%. Compared to previous search-in-reasoning approaches \cite{li2025search, islam2024open}, Agentic Reasoning demonstrates superior reasoning enhancement, outperforming Search-O1 by approximately 5\% overall. Furthermore, we show that this method is generally effective across different reasoning models, such as QwQ \cite{yang2024qwen2}, where it improves base model accuracy by over 10\%.

On GAIA (as shown in Table \ref{tab:gaia_results}), Agentic Reasoning establishes a new SOTA among all publicly available methods. Compared to OpenAI's Deep Research, which leverages its proprietary high-performance reasoning models, our approach surpasses it on Level 1 and Level 2 tasks while narrowing the gap to 2.26\% on Level 3. GAIA requires a combination of advanced reasoning, web browsing, and tool-use proficiency for successful completion. Our results demonstrate that Agentic Reasoning excels in handling complex tasks while maintaining strong generalization across diverse problem domains.

We also present a case study on a complex medical decision-making problem, as shown in Figure \ref{medex} The model autonomously executes code to compute the optimal $FiO_{2}$ (Fraction of Inspired Oxygen) for a patient, performs a web search to retrieve the most accurate PEEP (Positive End-Expiratory Pressure) value, and synthesizes both results to determine the best treatment plan.

\begin{table}[h]
    \centering
    \renewcommand{\arraystretch}{1.2}
    \resizebox{\hsize}{!}{
    \begin{tabular}{lcccc}
        \toprule
        \textbf{GAIA} & \textbf{Level 1} & \textbf{Level 2} & \textbf{Level 3} & \textbf{Avg.} \\
        \midrule
        Langfun  & 58.06 & 51.57 & 24.49 & 49.17 \\
        InspectReAct  & 67.92 & 59.30 & 30.77 & 57.58 \\
        h2oGPTe & 78.49 & 64.78 & 40.82 & 65.12 \\
        \hline
        \textbf{AgenticReasoning} & \textbf{74.36} & \textbf{69.21} & \textbf{45.46} & \textbf{66.13} \\
        \textit{Open AI Deep Research} $^\dagger$  & \textit{74.29} & \textit{69.06} & \textit{47.60} & \textit{67.36} \\
        \bottomrule
    \end{tabular}}
    \caption{Performance comparison on GAIA across different levels.}
    \label{tab:gaia_results}
\end{table}

\begin{table*}[h]
\centering
\begin{minipage}{0.45\linewidth}
    \centering
    \caption{Comparison with Human-Written Articles}
    \label{tab:comparison}
    \resizebox{\hsize}{!}{
    \begin{tabular}{lccc}
    \toprule
    & ROUGE-1 & ROUGE-L & Entity Recall \\
    \midrule
    Direct Gen & 27.32 & 13.13 & 6.11 \\
    RAG & 29.14 & 14.23 & 8.84 \\
    RAgent & 30.04 & 14.21 & 9.08 \\
    Search-O1 & 41.56 & 16.08 & 12.88 \\
    STORM & 47.93 & 17.42 & 15.43 \\
    \midrule
    \textbf{Ours} & \textbf{54.10} & \textbf{19.62} & \textbf{18.77} \\
    \bottomrule
    \end{tabular}}\label{tab:wiki}
\end{minipage}
\hfill
\begin{minipage}{0.45\linewidth}
    \centering
    \caption{Human Evaluation on Deep Research articles (Rate Range: 1-5). }
    \label{tab:rubric}
    \resizebox{\hsize}{!}{
    \begin{tabular}{lcccc}
    \toprule
    & Interest Level & Organization & Relevance & Coverage \\
    \midrule
    Direct Gen & 1.2 & 1.6 & 1.2 & 1.7 \\
    RAG & 1.4 & 2.1 & 1.9 & 2.3 \\
    RAgent & 1.6 & 2.3 & 1.6 & 2.6 \\
    Search-O1 & 2.5 & 2.8 & 2.1  & 3.2\\
    STORM & 2.9 & 3.2 & 2.9 & 3.7 \\
    Gemini-DR$^\dagger$ & 2.7 & 2.5 & 2.3 & 3.0 \\
    \midrule
    \textbf{Ours} & \textbf{3.7} & \textbf{4.6} & \textbf{4.2} & \textbf{4.1} \\
    \bottomrule
    \end{tabular}}\label{tab:human}
\end{minipage}
\end{table*}

\subsection{Deep Research}
We evaluate the deep research capability of our approach using the FreshWiki dataset \cite{shao2024assisting}, which curates high-quality, recent Wikipedia articles. The model is prompted directly with the topic and asked to generate the article. Evaluation covers article quality, assessed via ROUGE and entity recall. This task needs a comprehensive analysis of long-form generation while highlighting key challenges like bias transfer and factual consistency.

We also conduct an evaluation of Agentic Reasoning for deep research in open deep research tasks. A group of PhD-level experts in finance, medicine, and law were asked to formulate 15 to 30 professional research questions closely related to their respective fields. These questions were designed to require at least 20 minutes of in-depth research to answer comprehensively. There are in total 56 questions were collected. The experts would review the generated articles on interest level, organization, relevance, and coverage. More details are in the appendix.

We evaluate our method using the same underlying reasoning model, DeepSeek-R1, and compare it against various search-enhanced reasoning approaches, including RAG, RAgent \cite{islam2024open}, and Search-O1 \cite{li2025search}, as well as STORM \cite{shao2024assisting}, which employs a more complex agent-based workflow. Additionally, we benchmark our approach against the proprietary Gemini Deep Research \footnote{OpenAI Deep Research experiments are currently restricted by a high paywall.} on deep research tasks. As shown in Tables \ref{tab:wiki} and \ref{tab:human}, our results demonstrate that Agentic Reasoning consistently outperforms all RAG and search-based methods, as well as Gemini Deep Research, across all benchmarks. These findings highlight the effectiveness of structured reasoning and tool-augmented frameworks in enabling more advanced and efficient deep research.

\subsection{Analysis}
\subsubsection{Ablation on Toolbox}
\begin{figure*}[h]
    \begin{center}
    \includegraphics[width=0.95\textwidth]{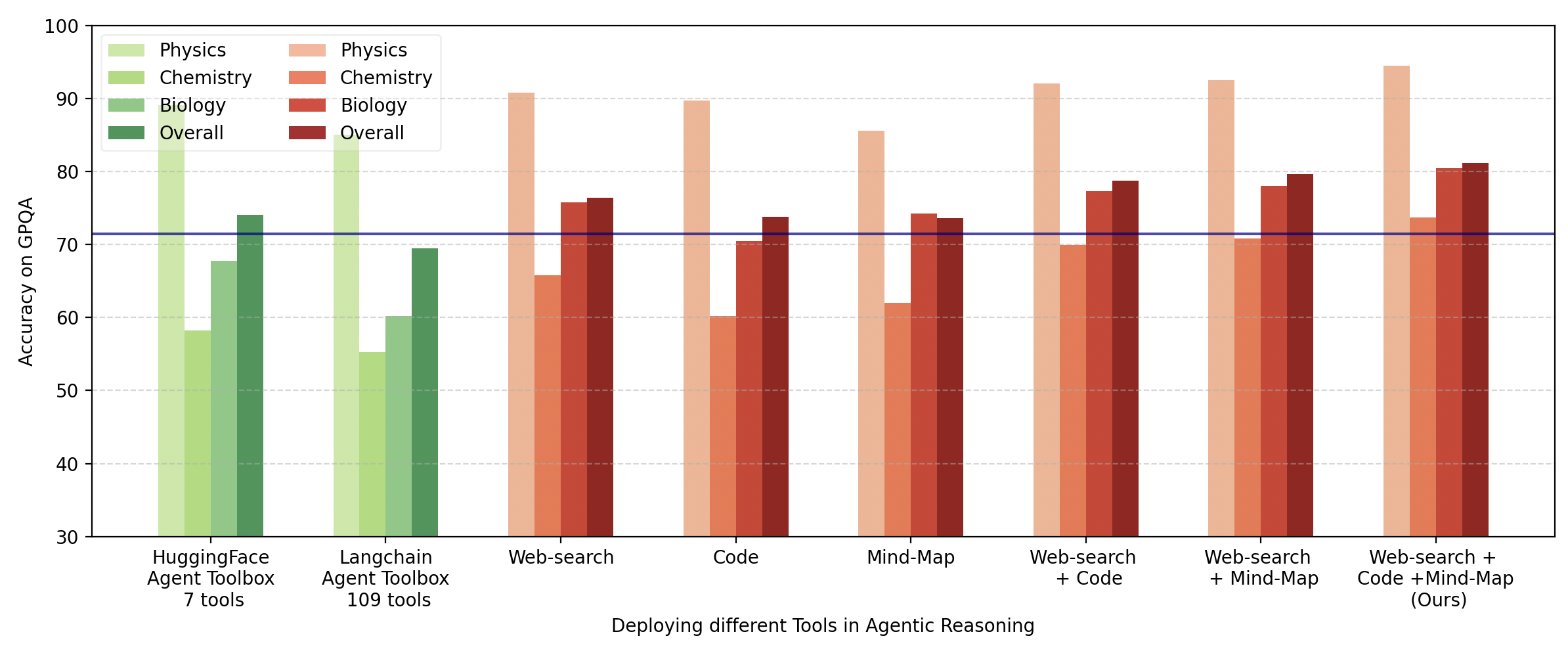}
    \end{center}
    \vspace{-10pt}
    \caption{The ablation study examines the impact of different tools in reasoning. Green ones represent external toolboxes, red ones are combinations of our proposed tools. The blue line is the overall performance of the base reasoning model. }\label{ablation}
\end{figure*}

We conduct experiments to explore the impact of integrating different tools in Agentic Reasoning and find that tool quality is far more important than quantity. Specifically, the combination of web search, coding, and Mind-Map agents proves to be the most effective across various tasks, including those requiring expert-level proficiency. As shown in Figure \ref{ablation}, we evaluated performance on GPQA using Hugging Face’s default agent toolbox with seven tools and LangChain with 109 tools. Surprisingly, adding more tools often degraded performance by increasing the risk of inappropriate tool selection. Many capabilities, such as translation or code interpretation, are already embedded within the reasoning model, making their external integration redundant. Moreover, inaccuracies in external tool outputs can negatively affect overall response quality.

Figure \ref{ablation} also presents an ablation study on the three proposed tools in this paper. We tested different tool combinations to assess their individual contributions to agentic reasoning. Among single-tool deployments, web search performed the best, while coding and Mind-Map achieved comparable results. Notably, combining tools yielded a synergistic effect—web search + Mind-Map or web search + coding provided greater improvements than the sum of their individual gains. The best performance was achieved when integrating all three: web search, Mind-Map, and coding.
\subsubsection{Ablation on Web-Search agent Design}
Integrating web search into LLMs has been widely explored in recent research \cite{li2025search, lewis2020retrieval, islam2024open}. In Agentic Reasoning, we investigate various web-search strategies to determine the most effective approach. Our ablation study primarily considers standard RAG and Knowledge Refinement, where retrieved sources are summarized for the response. Additionally, we incorporate Query Breakdown, Rerank, and Mind-Map Reasoning Context, key components in our Web-Search agent. Our findings reveal that Query Breakdown, Rerank, and Mind-Map Reasoning Context incrementally improve performance. Surprisingly, Knowledge Refinement, which is effective when used solely with RAG, becomes ineffective when combined with our three adopted components. This decline is primarily due to its redundancy with Rerank, which serves a similar role but proves more effective in most cases. As a result, our final Web-Search agent includes RAG, Query Breakdown, Rerank, and Mind-Map Reasoning Context for optimal performance.
\begin{table}[h]
    \centering
    \resizebox{\hsize}{!}{
     \begin{tabular}{l|cc|ccc|c}
        \hline
        & RAG & Knowledge & \textbf{Query} & \textbf{Rerank} & \textbf{Mind-Map} & \textbf{GAPA} \\
        &  & Refinement & \textbf{Breakdown} &  & \textbf{Reasoning Context} &  \\
        \hline
        Search-O1 & \checkmark & \checkmark &  &  &  & 74.6 \\
        Storm & \checkmark &  &  &  &  & 72.7 \\
        & \checkmark &  & \checkmark &  &  & 73.3 \\
        & \checkmark &  & \checkmark & \checkmark &  & 75.2 \\
        & \checkmark & \checkmark & \checkmark & \checkmark & \checkmark & 76.2 \\ 
        & \checkmark & \checkmark & \checkmark &  & \checkmark & 75.8 \\ 
        \hline
        \makecell{\textbf{Agentic} \\ \textbf{Reasoning}} & \checkmark &  & \checkmark & \checkmark & \checkmark & \textbf{76.4} \\
        \hline
    \end{tabular}}
    \caption{Comparison of different web-search approaches.}
    \label{tab:comparison}
\end{table}

\subsubsection{The Effect of Mind-Map}
We have shown in Figure \ref{ablation} that our quantitative results demonstrate that Mind-Map significantly enhances performance. In this section, we analyze its impact on reasoning in detail. Mind-Map proves particularly effective in maintaining long reasoning with tools and clarifying complex logical relationships.

We find that questions needs longer reasoning chains and more tool calls tend to be inherently more difficult, leading to lower accuracy, as shown in Figure \ref{mindmap}. Examining the model’s reasoning process, we find that LLMs often struggle to maintain coherence over long reasoning with massive tool calls. They may deviate from user queries, repeatedly call tools in the same way, or revisit previous errors, all of which negatively impact performance.

\begin{figure}[h]
    \begin{center}
    \includegraphics[width=0.48\textwidth]{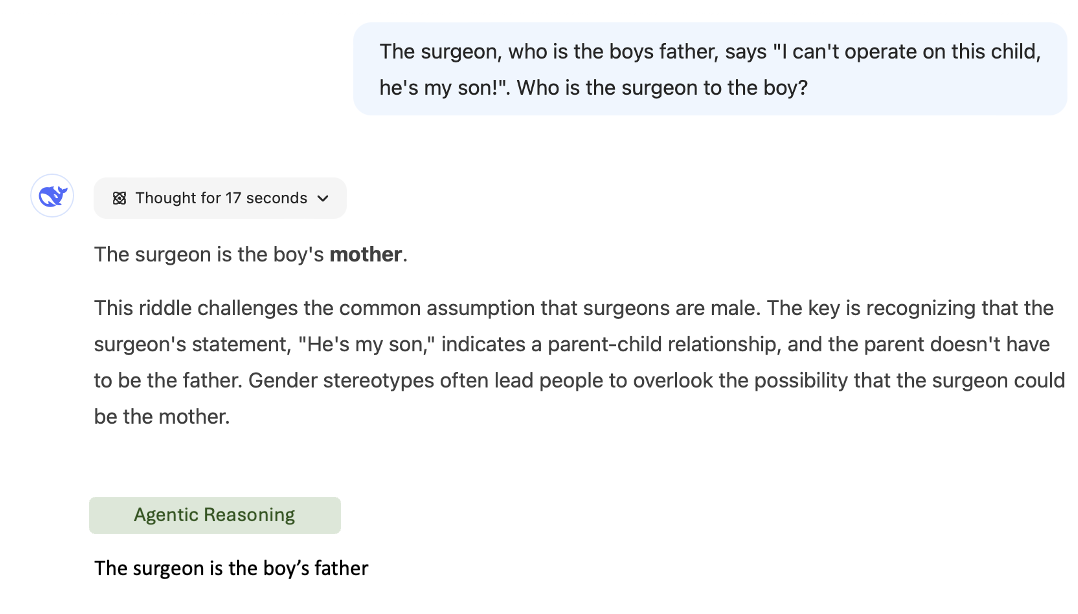}
    \end{center}
    \vspace{-10pt}
    \caption{A tricky question that misleads most LLMs is correctly answered by us.}\label{tricky}
\end{figure}

We introduced the Mind-Map agent to help the model manage its reasoning memory, ensuring coherent long reasoning and reducing errors. As shown in Figure \ref{mindmap}, this mechanism significantly improves performance, particularly on questions requiring long reasoning chains and more tool calls. The structured memory provided by the Mind-Map agent preserves prior reasoning steps, mitigating common pitfalls in extended reasoning tasks.

Mind-Map is also especially helpful for the tasks heavily rely on logic relationships. We find it helps to correctly answer tricky logic-based questions that frequently fool LLMs. A well-known example is a modified riddle: "The surgeon, who is the boy's father, says 'I can't operate on this child, he's my son!' Who is the surgeon to the boy?" As shown in Figure \ref{tricky}, DeepSeek-R1 took 17 seconds to process this question but still produced the wrong answer, a failure also observed in models from the GPT and Gemini series models. These models often fall for a political-correct corpus contaminated response, failing to recognize the obvious logical structure. However, in our Agentic Reasoning framework, the use of a Mind-Map allows the model to explicitly analyze the logical relationships between the entities [surgeon], [boy], and [father], leading to the correct answer.

This property also enables Mind-Maps to enhance deductive reasoning in strategic games. We tested our approach in Werewolf, a classic social deduction game where players assume hidden roles as either villagers or werewolves. Villagers aim to identify the werewolves through discussion, while werewolves deceive the group and eliminate players without being caught. To evaluate performance, we invited seven experienced Werewolf players, each with over five years of experience, to compete against our Agentic Reasoning model. The results show that our model achieved an impressive 72\% win rate, significantly surpassing both the expected statistical win rate and human performance in our experiment. In contrast, without Mind-Map, the model’s win rate dropped to 36\%. As the Mind-Map of the model’s reasoning process shown in Figure \ref{werewolf}, Mind-Map proved crucial in helping the model track relationships between players based on their spoken arguments. By maintaining a structured memory of interactions, it more effectively identified deception strategies, anticipated voting behaviors, and optimized its own disguise tactics. This result highlights that Mind-Map is not only a tool for structured logic but also a powerful enabler of strategic reasoning in dynamic, high-stakes environments.

\begin{figure}[h]
    \begin{center}
    \includegraphics[width=0.45\textwidth]{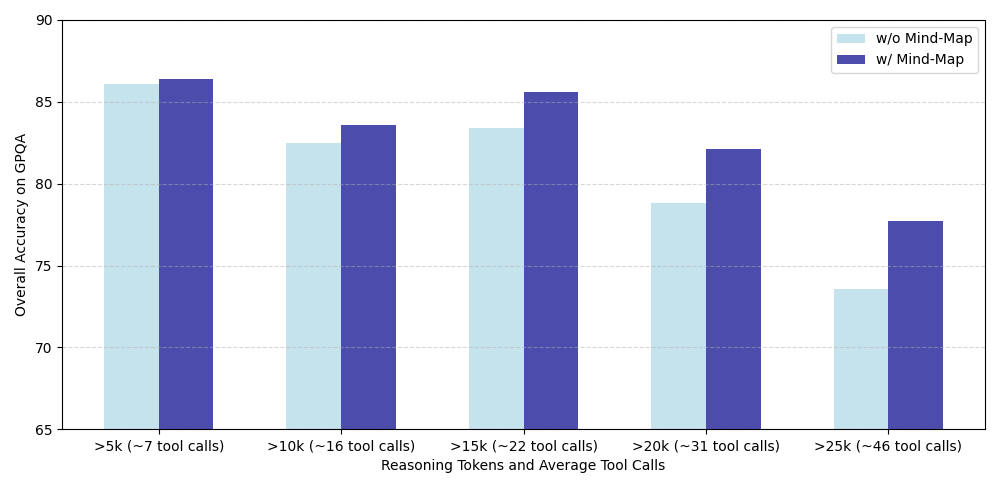}
    \end{center}
    \vspace{-10pt}
    \caption{Mind-Map improves performance on questions need long reasoning.}\label{mindmap}
\end{figure}

\begin{figure}[h]
    \begin{center}
    \includegraphics[width=0.45\textwidth]{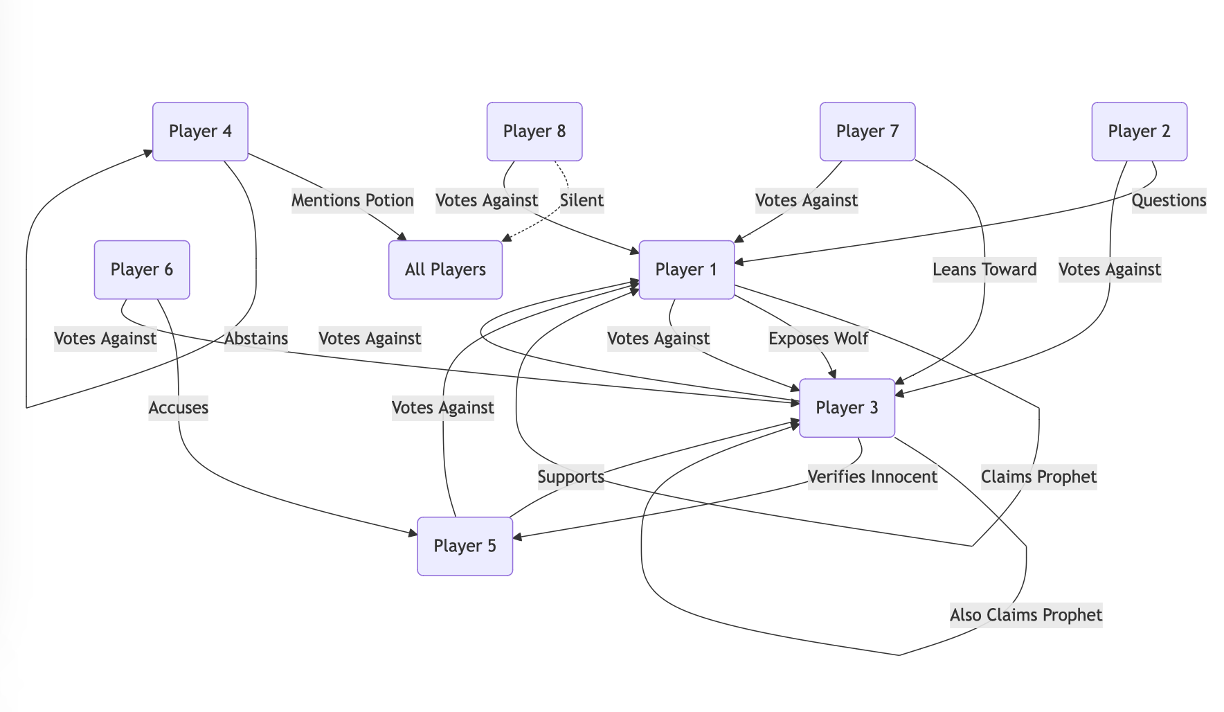}
    \end{center}
    \vspace{-10pt}
    \caption{Mind-Map in playing werewolf game. The first round and the second round. Player8 is the model.}\label{werewolf}
\end{figure}

\section{Related Work}
The concept of multi-agent collaboration in LLMs has gained attention with frameworks like AutoGPT \cite{yang2023auto} and LangChain Agents \cite{pandya2023automating}, allowing models to interact with external APIs, search engines, and computational environments. While these frameworks introduce modular reasoning, they often lack optimized task delegation and structured integration, reducing their effectiveness in long-chain reasoning tasks. Recent research on Hierarchical Planning with LLMs \cite{luo2023obtaining} and Task-Specific AI Agents \cite{wu2024medical} explores structured agent cooperation for problem-solving. However, these approaches still do not deeply integrate agent tools within reasoning chains and fail to systematically explore optimal agent combinations that maximize reasoning performance.

Previous studies focus a lot on integrating the search capability into LLMs. Recent agentic RAG systems\cite{khaliq2024ragar, islam2024open, li2025search} have enabled models to autonomously determine when and what knowledge to retrieve, enhancing their planning and problem-solving capabilities. Additionally, research has explored \cite{li2025search} integrating Web-Search agent into the reasoning model, like QwQ \cite{teamqwq} demonstrating the potential of search augmentation in structured reasoning. However, existing approaches have primarily focused on single-agent enhancements, neglecting the potential synergy of multiple agentic tools. Moreover, prior works have yet to integrate such tools with state-of-the-art reasoning models like DeepSeek-R1 or OpenAI-O1, limiting their effectiveness in solving highly complex tasks.

\section{Conclusion}
We introduced Agentic Reasoning, a framework that enhances LLM reasoning by integrating Mind-Map, web search, and coding. Our approach improves problem-solving and deep research capabilities, outperforming existing models in expert-level QA and real-world research tasks. Agentic Reasoning outperforms existing methods in both quantitative benchmarks and human evaluations. Future work will explore task-specific tools integration and test-time computing to further enhance AI’s reasoning capabilities.
\clearpage
\section{Limitations}

Despite the strong performance of Agentic Reasoning, several limitations remain that warrant further research and refinement.

\paragraph{Computational Overhead and Efficiency.} 
Integrating multiple external agents, including web search, Mind-Map, and code execution, significantly increases computational costs and inference latency. While these components enhance reasoning depth, their sequential invocation introduces bottlenecks, limiting real-time applicability. Future work would explore techniques such as agent prioritization, caching strategies, or adaptive invocation mechanisms to optimize efficiency without sacrificing accuracy.

\paragraph{Reliance on External Knowledge Sources.} 
The effectiveness of Agentic Reasoning depends on the quality of retrieved knowledge, particularly in web search. The system lacks built-in verification mechanisms to assess the credibility of sources, making it susceptible to misinformation or biased content. Developing trust-aware retrieval mechanisms, such as fact-checking agents or weighted source reliability scores, could mitigate this risk and improve robustness in knowledge-intensive domains.

\paragraph{Interpretability and Trustworthiness.}  
While the Mind-Map agent provides structured reasoning memory, the overall decision-making process remains highly dependent on LLMs. This reliance introduces the risk of hallucinations, which can derail the entire reasoning process, especially in complex, multi-step tasks. In high-stakes domains such as medical AI or legal reasoning, even minor inaccuracies can lead to significant consequences. Ensuring reliability requires additional safeguards, such as fact-verification mechanisms, confidence estimation, or human-in-the-loop oversight, to mitigate the risks associated with LLM-driven reasoning.

\section*{Acknowledgments}
Junde Wu is supported by the Engineering and Physical Sciences Research Council (EPSRC) under grant EP/S024093/1 and GE HealthCare. Jiayuan Zhu is supported by the Engineering and Physical Sciences Research Council (EPSRC) under grant EP/S024093/1 and Global Health R\&D of Merck Healthcare, Ares Trading S.A. (an affiliate of Merck KGaA, Darmstadt, Germany), Eysins, Switzerland (Crossref Funder ID: 10.13039/100009945). Yueming Jin is supported by the Ministry of Education Tier 1 grant, NUS, Singapore (24-1250-P0001).

\bibliography{acl_latex}
\clearpage

\section{Appendix}

\subsection{Human Evaluation Survey}
Please assess each response generated by the model based on the following criteria. Provide your rating on a scale from 1 to 5, where 1 is the lowest and 5 is the highest. You may also leave optional comments to clarify your reasoning.

\begin{enumerate}
    \item \textbf{Interest Level (Int.)}
    \begin{itemize}
        \item How engaging and intellectually stimulating is the generated response?
        \item Rating Scale: 
        \textbf{1:} Not engaging - fails to capture interest. \textbf{2:} Somewhat uninteresting - lacks depth or novelty. \textbf{3:} Neutral - informative but not particularly engaging. \textbf{4:} Engaging - provides depth and insight. \textbf{5:} Highly engaging - deep and thought-provoking.
        \item \textbf{Optional Comment}: What aspects of the response contributed to or detracted from its interest level?
    \end{itemize}
    
    \item \textbf{Organization (Org.)}
    \begin{itemize}
        \item How well-structured and logically organized is the response?
        \item Rating Scale: 
        \textbf{1:} Very disorganized - hard to follow. \textbf{2:} Somewhat disorganized - requires effort to understand. \textbf{3:} Neutral - moderately structured but could be clearer. \textbf{4:} Well-organized - logical and easy to follow. \textbf{5:} Exceptionally structured - very clear and logically ordered.
        \item \textbf{Optional Comment}: Are there any areas where the response could be better structured?
    \end{itemize}
    
    \item \textbf{Relevance (Rel.)}
    \begin{itemize}
        \item How relevant is the response to the research question posed?
        \item Rating Scale: 
        \textbf{1:} Not relevant - off-topic or misleading. \textbf{2:} Somewhat relevant - partially addresses the question. \textbf{3:} Neutral - addresses the question but with some tangents. \textbf{4:} Mostly relevant - minor deviations but generally on point. \textbf{5:} Highly relevant - fully addresses the question.
        \item \textbf{Optional Comment}: Did the response stay on topic? If not, how did it deviate?
    \end{itemize}
    
    \item \textbf{Coverage (Cov.)}
    \begin{itemize}
        \item How comprehensively does the response cover the question?
        \item Rating Scale: 
        \textbf{1:} Superficial - lacks depth and critical information. \textbf{2:} Somewhat incomplete - covers only basic aspects. \textbf{3:} Neutral - adequate coverage but missing key details. \textbf{4:} Mostly complete - only minor gaps. \textbf{5:} Fully comprehensive - deeply covers all necessary aspects.
        \item \textbf{Optional Comment}: Are there any areas where additional information would improve the response?
    \end{itemize}
\end{enumerate}

Thank you for your participation!

\subsection{Additional Experiments}
\subsubsection{Ablation Study on Memory Strategies}
We have conducted a comparison of several alternative memory strategies within our agentic reasoning framework to replace the Mind-Map module (Table \ref{supp:7.2.1}). These include: no memory (None Mem), using raw reasoning content as memory (Raw Mem), as well as integrating existing methods such as Read-Agent \cite{lee2024human}, MemoryBank \cite{zhong2024memorybank}, and MemGPT \cite{packer2023memgpt}. We evaluated all approaches on the GAIA benchmark and found that our Mind-Map strategy consistently achieved the highest performance across all settings. 

\begin{table}[ht]
\centering
\caption{Compare to alternative memory strategies.}
\vspace{-10pt}
\resizebox{0.48\textwidth}{!}{
\centering
\begin{tabular}{c|ccc|c}
                & Level 1        & Level 2        & Level 3        & Avg.           \\ \hline
None Mem        & 62.37          & 46.54          & 24.49          & 46.18          \\
Raw Mem         & 62.37          & 47.80          & 26.53          & 47.84          \\
Read-agent    & 64.28          & 51.57          & 27.70          & 49.83          \\
MemoryBank  & 68.41          & 55.18          & 32.65          & 53.49          \\
MemGPT         & 72.04          & 66.70          & 42.11          & 65.12          \\
Mind-Map (ours) & \textbf{74.36} & \textbf{69.21} & \textbf{45.46} & \textbf{66.13}
\end{tabular}}
\label{supp:7.2.1}
\vspace{-10pt}
\end{table}

\begin{table*}[ht]
\centering
\caption{Compare performance and time consumed on 56 deep research questions.}
\vspace{-10pt}
\resizebox{0.98\textwidth}{!}{
\centering
\begin{tabular}{c|ccccc}
                                                                                                    & Avg. time consumed / per question & Interest Level & Organization & Relevance & Coverage \\ \hline
Perplexity-deep research                                                                            & 3.1 mins                          & 2.0            & 2.1          & 1.6       & 2.1      \\
Gemini-deep research                                                                                & 7.7 mins                          & 2.7            & 2.5          & 2.3       & 3.0      \\
Ours                                                                                                & 6.8 mins                          & 3.7            & 4.6          & 4.2       & 4.1      \\
GPT-deep research                                                                                   & 17.8 mins                         & 4.1            & 4.8          & 4.2       & 4.5      \\
\begin{tabular}[c]{@{}c@{}}Human w/ DeepSeek R1,\\ Cursor and Web-Search (6 questions)\end{tabular} & 1h 48 mins                        & 4.5            & 4.8          & 4.8       & 4.5     
\end{tabular}}
\label{supp:7.2.2}
\vspace{-10pt}
\end{table*}

\subsubsection{Efficiency Analysis}
We conducted a comparison on 56 deep research questions, measuring both performance and time consumed. We compared our model against three related proprietary systems: Perplexity-Deep Research, Gemini-Deep Research, and GPT-Deep Research. As shown in Table \ref{supp:7.2.2}, we can see our model achieves substantially better performance than Perplexity-Deep Research while being faster, and it outperforms Gemini-Deep Research with comparable latency. While slightly behind GPT-Deep Research in performance, our model runs significantly faster, and we believe the performance gap is likely attributable to GPT-Deep Research being built on a more advanced underlying reasoning model.

\subsubsection{Websearch and Coding as Agent}
Using agentic tool calls rather than direct API calls offers several advantages:

\begin{itemize}
\item[1.] \textbf{Overcoming single-model token limitations}: By structuring the workflow agentically, the system can break free from the token generation limits of a single LLM. This enables it to produce longer, higher-quality reasoning chains than would be possible within the token budget of a single model call.

\item[2.] \textbf{Managing uncertainty and reducing error propagation}: Agents can self-monitor and assign varying levels of confidence to their outputs. As noted earlier, this mechanism helps the reasoning model treat low-confidence outputs as tentative, thereby mitigating cascading errors across components.
\end{itemize}

For example, in the web-search agent, if insufficient information is retrieved to confidently answer a query, the agent may respond with something like: \textit{``Given the limited knowledge retrieved, a possible answer might be... However, due to the lack of sufficient source information, additional data is needed to provide a more accurate response.''}

This uncertainty is explicitly communicated back to the reasoning model, allowing it to treat the response as tentative, rather than relying on it as a final answer. In our experiments, we found that this self-awareness and feedback mechanism helps reduce cascading errors and improves overall robustness. We support this claim with an ablation study on the GAIA benchmark, comparing our agent-based system with a version that uses direct API calls for coding and web-search. Table \ref{supp:7.2.3} shows that the agent-based design significantly reduces errors and improves performance, especially on level-3 hard problems, validating our approach.

\begin{table}[]
\centering
\caption{Compare with a version that uses direct API calls for coding and web-search.}
\vspace{-10pt}
\resizebox{0.48\textwidth}{!}{
\centering
\begin{tabular}{c|ccc|c}
              & Level 1 & Level 2 & Level 3 & Avg.  \\ \hline
API Calling   & 60.22   & 46.54   & 24.49   & 47.18 \\
Agentic Tools & \textbf{74.36}   & \textbf{69.21}   & \textbf{45.46}   & \textbf{66.13}
\end{tabular}}
\label{supp:7.2.3}
\vspace{-10pt}
\end{table}

\begin{itemize}
\item[3.] \textbf{Task-specific model modularity}: The agentic design allows us to assign different LLMs to different tasks. For example, Claude-Sonnet tends to perform better on coding tasks, so we route the coding agent to use it specifically. Similarly, for tasks like summarizing web search results, we can use a lightweight, non-reasoning model to preserve efficiency. This modular setup allows for both improved performance and optimized resource usage by matching the best model to each subtask.
\end{itemize}

\end{document}